%% file: main.tex
\documentclass[10pt,twocolumn,letterpaper]{article}

\usepackage{cvpr}              

\input{preamble}

\definecolor{cvprblue}{rgb}{0.21,0.49,0.74}
\usepackage[pagebackref,breaklinks,colorlinks,allcolors=cvprblue]{hyperref}

\def\paperID{1625} 
\def\confName{CVPR}
\def\confYear{2025}

\title{Exploring Timeline Control for Facial Motion Generation}

\author{Yifeng Ma\textsuperscript{\rm 1} , Jinwei Qi\textsuperscript{\rm 2} , Chaonan Ji\textsuperscript{\rm 2} , Peng Zhang\textsuperscript{\rm 2} , Bang Zhang\textsuperscript{\rm 2} , Zhidong Deng\textsuperscript{\rm 1} , Liefeng Bo\textsuperscript{\rm 2} \\
\textsuperscript{\rm 1} Department of Computer Science and Technology, Tsinghua University \\
\textsuperscript{\rm 2} Tongyi Lab, Alibaba Group
}

\begin{document}

\twocolumn[{%
\maketitle
\begin{figure}[H]
\hsize=\textwidth %
\centering
\includegraphics[width=\textwidth]{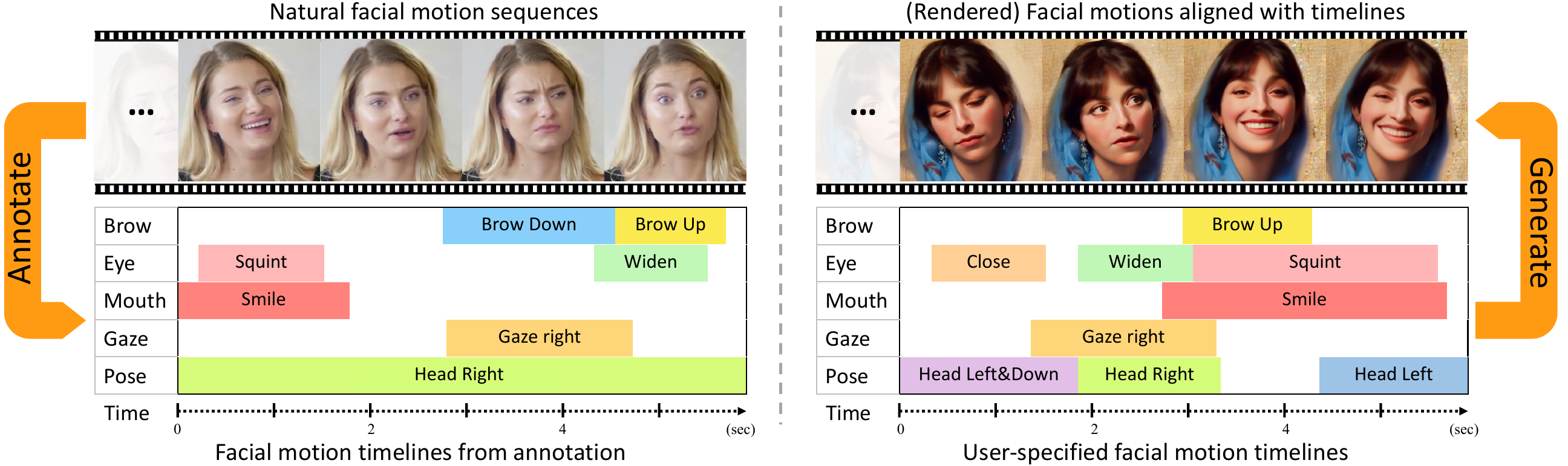}
\caption{
We introduce a new control signal for facial motion generation: \textbf{timeline control}. We first utilize a labor-efficient approach to annotate the time intervals of facial motion at a frame-level granularity. Using the annotations, we propose a model that can generate natural facial motions aligned with an input timeline. Compared to previous controls like audio and text, timeline control enables precise temporal control of facial motions. In this paper, facial motions are rendered into photorealistic videos for better visualization.
} 
\label{fig:teaser}
\end{figure}
}]

\maketitle

\input{sec/0_abstract}    
\input{sec/1_intro}
\input{sec/2_related_work}
\input{sec/3_method}

\input{sec/4_experiment}
\input{sec/5_conclusion}

{
    \small
    \bibliographystyle{ieeenat_fullname}
    \bibliography{main}
}


\end{document}

%% file: preamble.tex
\usepackage{graphicx}
\usepackage{tabularx}
\usepackage{makecell}
\usepackage{subcaption}
\usepackage{multirow}
\usepackage{amsmath}
\usepackage{amssymb}
\usepackage{mathrsfs}
\usepackage{bm}

\usepackage{float}

\newcommand{\supp}{\textit{Supplementary Material}\xspace}
\newcommand{\suppvideo}{\textit{Supp. Video}\xspace}

\newcommand{\vect}[1]{\bm{#1}}
\newcommand{\eqword}[1]{{\text{#1}}}

\newlength\savewidth\newcommand\shline{\noalign{\global\savewidth\arrayrulewidth
  \global\arrayrulewidth 1pt}\hline\noalign{\global\arrayrulewidth\savewidth}}

%% file: sec/0_abstract.tex
\begin{abstract}
This paper introduces a new control signal for facial motion generation: timeline control. 
Compared to audio and text signals, timelines provide more fine-grained control, such as generating specific facial motions with precise timing.
Users can specify a multi-track timeline of facial actions arranged in temporal intervals, allowing precise control over the timing of each action.
To model the timeline control capability, We first annotate the time intervals of facial actions in natural facial motion sequences at a frame-level granularity. 
This process is facilitated by Toeplitz Inverse Covariance-based Clustering to minimize human labor.
Based on the annotations, we propose a diffusion-based generation model capable of generating facial motions that are natural and accurately aligned with input timelines.
Our method supports text-guided motion generation by using ChatGPT to convert text into timelines.
Experimental results show that our method can annotate facial action intervals with satisfactory accuracy, and produces natural facial motions accurately aligned with timelines.
\end{abstract}

%% file: sec/1_intro.tex
\section{Introduction}
Generating vivid facial motions has drawn growing attention due to its broad applications, including digital human generation and filmmaking. 
To produce desired facial motions, current methods use audio or text to provide guidance.
However, these methods lack a critical ability that users often require: generating specific facial motions with precise timing.
For example, users may want to generate a brow raise between frames 10 and 30 while simultaneously generating a smile between frames 14 and 43.
Audio-driven methods~\cite{tian2024emo, xu2024vasa} can only generate motions synchronized with the audio. 
Text signals offer only coarse-grained descriptions of facial motions and lack frame-level detail. Text-driven approaches~\cite{wang2023agentavatar, wang2024instructavatar, wu2024mmhead, brooks2024video} rely on temporal adverbial cues (e.g., \emph{then}) for rough timing guidance.

To achieve more fine-grained control, we introduce a novel control signal: timeline control for facial motion generation. 
In this setup, users can generate facial motions by inputting an intuitive timeline that contains several temporal intervals, each corresponding to a desired facial action. This setup enables users to manage the timing of each action.

Achieving timeline control is very challenging. 
Frame-level control of facial motions requires the model to achieve exceptional precision in generating accurate actions.
To model timeline control capability, it is crucial to annotate the precise start and end frames of facial actions, a challenge that existing methods have yet to overcome.
Existing methods~\cite{wang2023agentavatar, wu2024mmhead} represent facial motion sequences as a time series of motion descriptors, such as Facial Action Units (AUs)~\cite{ekman2002facial} or blendshapes~\cite{wang2022faceverse},
and then leverage ChatGPT to summarize these time series for annotation. However, ChatGPT's limited sensitivity to the temporal dynamics of facial motions prevents it from determining the exact start and end frames of facial actions. 
Another annotation approach is using thresholds, like labeling the eye as "closed" if the blendshape \emph{eyeBlink} exceeds 0.4. However, setting thresholds for complex actions, such as brow motions, is challenging. Additionally, some actions are determined by the relative values of multiple motion descriptors rather than a single one.


To address this issue, we adopt Toeplitz Inverse Covariance-based Clustering (TICC)~\cite{hallac2017toeplitz} to annotate the start and end frames of facial actions. 
TICC can segment the time series into a sequence of intervals, with each interval containing a single motion pattern. Each interval has clearly defined start and end frames. 
Once the start and end frames of the intervals are identified, the remaining task is to determine the facial action represented by each interval. This can be addressed by another feature of TICC: its ability to group the segmented intervals into several clusters based on the similarity of their motion patterns, with each cluster containing intervals that exhibit similar motion patterns.
The automatic clustering process eliminate the need for manual threshold setting and considers the relationships between multiple motion descriptors.
Once these clusters are obtained, we can determine the overall action (\eg significantly raised brows) represented by each cluster by analyzing a few representative patterns within it. 
This process requires minimal labor.
Using this procedure, we analyze motions in different facial regions individually and achieve frame-level annotation for the brows, eyes, mouth, gaze, and head motions in a labor-efficient manner.

Using frame-level facial motion annotations, we propose a diffusion-based generation model that produces natural facial motions accurately aligned with the input timeline.
The model generates motions for each facial region (\eg upper face, lower face) separately. This decoupling can help address the adverse impact of motion coupling in the data, thereby improving accuracy.
However, some couplings are necessary for motion naturalness, such as a slight head lift during a brow raise.
To properly manage coupling, the generation model is divided into a base network and multiple branch networks, with each branch network dedicated to a facial region.
The base network encodes global facial motion couplings into base features.
The branch network takes base features and the timeline of the corresponding region as input to accurately generate facial motions for that region while maintaining natural couplings.
We use FaceVerse\cite{wang2022faceverse} 3DMM coefficients to represent facial motions for generation.
Since photorealistic facial motion enables more accurate assessments of motion realism  than mesh-based motion~\cite{ng2024audio2photoreal}, we render motions into photorealistic videos by a diffusion-based renderer.
Our method enables text-guided facial motion generation by using ChatGPT to convert text into timelines. ChatGPT effectively translates simple natural language descriptions into timelines, which our model transforms into facial motions. Users can modify these timelines, allowing precise timing control for text-driven applications.






In summary, our contributions are as follows:

\begin{itemize}
    \item We are the first to develop a labor-efficient approach to annotate temporal intervals of facial actions. 
    Such fine-grained annotations enable precise modeling of the temporal dynamics of facial motions.
    \item We are the first to achieve timeline control for facial motion generation. This enables users to generate specific facial motions with precise timing. Our method also supports generating actions from natural language text.
    \item Detailed evaluations show that our method generates accurate motion annotations and produces natural, precise facial motions from timeline inputs. 
    
\end{itemize}




\begin{figure*}
  \centering
  \includegraphics[width=0.98\linewidth]{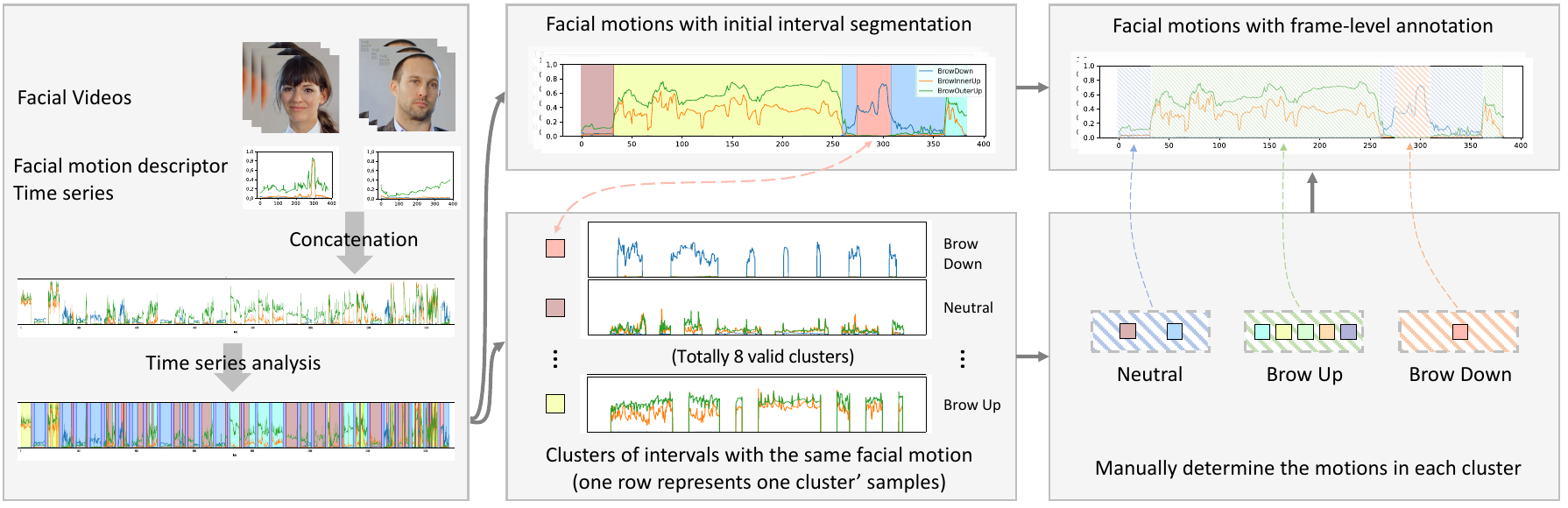}
  \vspace{-3mm}
  \caption{\textbf{The pipeline of frame-level facial motion annotation} (using brow motions as an example). We first extract facial motion descriptors (blendshapes)  from natural facial motion videos and concatenate the results to create a facial motion time series for time series analysis. This analysis can simultaneously segment the sequence into a series of motion patterns and cluster similar patterns, resulting in multiple clusters, each containing consistent facial motion patterns. Then, by inspecting a few patterns, we identify the facial motions each cluster represents, thereby obtaining frame-level facial motion annotations for all videos. }
  \vspace{-4mm}
  \label{fig:annotation_pipeline}
\end{figure*}

%% file: sec/2_related_work.tex
\section{Related Work}

\noindent\textbf{Facial Motion Annotation.}
Compared to human motion annotation~\cite{plappert2016kit,guo2022generating,mahmood2019amass,guo2020action2motion}, facial motion annotation is in its early stages.
Existing methods~\cite{wu2024mmhead,wang2024instructavatar,ma2023talkclip,wang2023agentavatar,yu2022celebvtext} generate annotations that either overlook temporal changes or describe them only roughly with temporal adverbs.
Some datasets~\cite{yap2020samm,7820164} label intervals of macro/micro expressions rather than facial actions used for facial motion generation. These datasets also require intensive manual labeling.
Facial expression spotting methods~\cite{deng2024spotformer,yap20223d,wang2021mesnet,yu2022facial,guo2023micro} are also restricted to label macro/micro-expressions.
In this paper, we aim to labor-efficiently generate frame-level facial action annotations.

\noindent\textbf{Facial Motion Generation.}
Existing methods use audio or text as the control signal and cannot generate specific motions at precise timing. Audio-driven methods (2D~\cite{tian2024emo,xu2024vasa,jiang2024loopy,guan2024resyncer,chen2024echomimic,xu2024hallo,wei2024aniportrait,tan2025edtalk,zhang2024personatalk,ye2023geneface,li2025talkinggaussian,nocentini2024scantalk,he2023gaia,zhang2023sadtalker,wang2024styletalk++,prajwal2020lip} and 3D methods~\cite{zhao2024media2face,cudeiro2019capture,li2024kmtalk,peng2023emotalk,aneja2024facetalk,danvevcek2023emotional,fan2024unitalker,yang2024probabilistic,sun2024diffposetalk,xing2023codetalker,fan2022faceformer,karras2017audio}) can only generate facial motions synchronized with the audio.
Text-driven methods~\cite{wang2023agentavatar,wang2024instructavatar,ma2023talkclip,wu2024mmhead,brooks2024video,ling2024posetalk} can only use temporal adverbial phrases to control timing roughly.
Rule-based methods~\cite{cassell1994animated,pelachaud1996generating} offer temporal control but often produce unnatural motions due to deviations from real movement distributions.
This paper introduces timeline control to generate natural facial motions with precise timing.

Timeline control has been utilized in human motion generation~\cite{petrovich2024multi,athanasiou2022teach,barquero2024seamless,shafir2023human} by first generating motion segments and then piecing them together. However, due to the rapid changes in facial movements, this approach is unsuitable, necessitating a new model structure for effective generation.


%% file: sec/3_method.tex
\section{Method}

In this work, we first annotate the temporal intervals of facial action for natural facial motion sequences. Based on these annotations, we develop a generation model that generates natural facial motions that are accurately aligned with an input timeline.
Our method also supports text-guided motion generation by leveraging ChatGPT to convert text into timelines.


\subsection{Annotating temporal intervals of facial actions}
\label{sec:method_facial_motion_annotation}

\begin{table}[t]
\centering
\setlength{\tabcolsep}{0.5mm}{
\begin{tabular}{c|c}
\toprule  
Facial Region \ \  & Motion Categories  \\
\shline
Brow   & BrowUp, BrowDown, Neutral  \\
Eye  & EyeSquint, EyeWiden, EyeClose, Neutral \\
Mouth  & SoftSmile, Smile, MouthFrown, Neutral \\
Gaze  & Left, Right, Up, Down, Neutral \\
Head  & Left, Right, Up, Down, Neutral \\

\bottomrule 
\end{tabular}}
\vspace{-3mm}
\caption{Facial actions annotated by our method.}
\vspace{-4mm}
\label{table:facial_motion_category}
\end{table}

\noindent\textbf{Problem Formulation.} For each video $\vect{V}$ in dataset, our method aims to generate frame-level facial motion annotations $\vect{A}=[\vect{a}_i]_{i=1}^{L}$  for each video frame $[\vect{v}_i]_{i=1}^{L}$. \cref{table:facial_motion_category} shows the annotated facial actions.
The annotation $\vect{a}_i$ for each frame is a vector of binary values, with each dimension representing a non-neutral facial action. A dimension is set to 1 if the action occurs, otherwise, it is 0.


\noindent\textbf{Annotation Process Overview.}
We achieve the annotation through time series analysis TICC. \cref{fig:annotation_pipeline} shows the pipeline:
(1) We extract proper motion descriptors for each facial region to obtain facial motion time series.
(2) We apply TICC to simultaneously segment the time series into a sequence of motion patterns and cluster similar patterns. This results in several clusters of intervals, each containing similar facial motions.
(3) Finally, for each cluster, we can infer the actions of all intervals by examining only a few intervals. Once the action for each cluster is identified, the actions for all intervals are determined, resulting in facial action annotations at a frame-level granularity.

\noindent\textbf{Facial Motion Descriptor.}
We observe that previously used AUs
lack sufficient precision, while ARKit blendshapes offer a more accurate representation. The publicly available MediaPipe blendshape detector struggles to detect certain blendshapes (\eg eyeWide, mouthFrown), so we use an in-house blendshape detector.
ARKit blendshapes are a set of coefficients that describe facial expressions, with each coefficient representing the movement of a specific facial region. Each coefficient ranges from 0 to 1, indicating the motion intensity.

We use the blendshapes corresponding to each facial region to construct motion time series for each region, resulting in time series data for the eye, brow, and mouth areas. 
For the eye region, we selected \emph{eyeBlink, eyeSquint, eyeWide}, and for the brow region, we selected \emph{browDown, browInnerUp, browOuterUp}.
For the mouth region, we use only expression-related blendshapes (\emph{mouthSmile}, \emph{mouthStretch}, \emph{mouthFrown}) to avoid interference from speech-related movements.
ARKit blendshapes provide coefficients for the left and right sides. As facial motions in our dataset are largely symmetrical, we use only the left-side coefficients to simplify analysis.

For gaze and head pose motion, we use a 3DMM model FaceVerse~\cite{wang2022faceverse}  to detect coefficients. For gaze, we select eye coefficients, for head pose, we select angle coefficients.

\begin{figure*}
  \centering
  \includegraphics[width=0.98\linewidth]{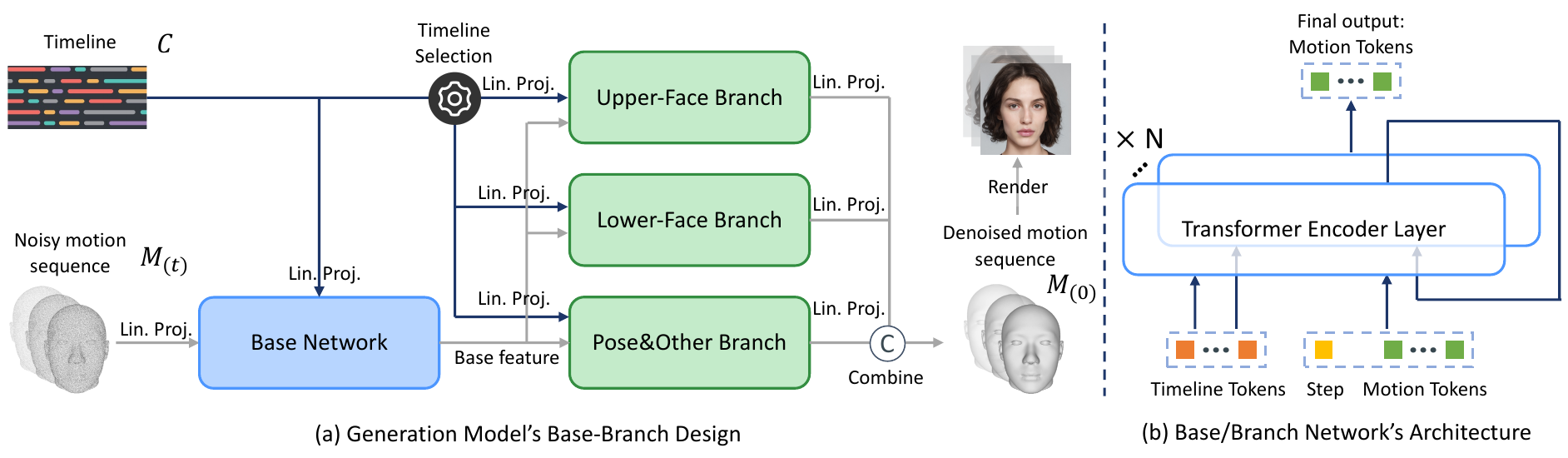}
  \vspace{-3mm}
  \caption{\textbf{Illustration of generation model.} (a) \textbf{Base-Branch Design.}  The base network takes the timelines of all facial regions as input and outputs base features that model the global facial motion couplings. Through timeline selection, each region's timeline is directed to its respective branch network. Since head pose is interconnected with all facial movements, the pose branch receives timelines of all regions. Each branch network takes the timeline of its corresponding region to generate the facial motions for that region. These motions are then combined to produce the overall motion of the entire face.
  \emph{Lin. Proj.} denotes \emph{Linear Projection}.
  (b) \textbf{Base/Branch Network's Architecture.} 
  Timeline control guides motion generation through cross-attention. The initial timeline tokens remain unchanged and are added at each layer. For clarity, the diffusion step (omitted in sub-figure (a) for clarity) is applied to each base and branch network.
  }
  \vspace{-4mm}
  \label{fig:generation_pipeline}
\end{figure*}

\noindent\textbf{Analyzing Facial Motion Time Series.}
We utilize TICC~\cite{hallac2017toeplitz} for time series analysis. 
 TICC is designed to handle a single long sequence (CubeMarker~\cite{honda2019multi} can analyze multiple time series analysis but performs poorly.). However, our data consists of multiple short sequences. To address this issue, we concatenate all the short sequences into a single long sequence for analysis. To avoid the impact of abrupt transitions between different videos, we introduce a "null sequence" between videos to separate them. The null sequence is implemented as a length-100 sequence with values set to -1. These null sequences will eventually be grouped into a single cluster, which will not affect others. 




After the time series analysis, the video is annotated with a sequence of motion pattern intervals and similar patterns are clustered into a few clusters. 
Therefore, for each cluster, we inspect a few intervals to determine the facial action they represent. This facial action is then considered to be the action for all intervals within that cluster.
Different clusters may represent variations of the same facial action category, such as a high eyebrow raise versus a moderate one.
We categorized the clusters into facial motion categories, such as eyebrow raise, frown, and neutral, to obtain frame-level annotations.

We observe that eye closure frequently produces sharp spikes in the time series, which disrupts the analysis of squinting and widening. Therefore, we analyze eye closure and squinting/widening separately.
We also find that threshold-based annotation, which classifies actions using predefined thresholds (e.g., labeling the eye as "closed" if \emph{eyeBlink} exceeds 0.4), achieves high accuracy for eye closure, gaze, and head pose. Consequently, we use this method for these regions, while applying time series analysis for others.


\noindent\textbf{Detect Facial Motion for Unseen Videos.} 
After learning various facial motion patterns from the data, TICC can also be applied to previously unseen facial motion sequences and detect facial action intervals. We use this function to assess the accuracy of facial motion generation in our evaluation.

\subsection{Facial Motion Generation from Timelines}

\noindent\textbf{Problem Formulation.} 
Given a timeline control $\vect{C}=[\vect{c}_i]_{i=1}^{L}$ for each video frame, the generation model $\mathcal{G}$ aims to generate natural facial motions $\vect{M}=[\vect{m}_l]_{l=1}^{L}$ that align with the timeline control. The control for each frame $\vect{c}_i$ follows the same format as the annotation. Generated facial motions are rendered by a diffusion-based renderer into photorealistic videos for better visualization.

\noindent\textbf{Discussion: The coupling of facial motions and its impact on motion accuracy and naturalness.}
\label{sec:facial_motion_generation}
Facial motions across different regions are coupled. For example, when people smile, they might also squint and lower their eyebrows. These couplings can reduce the precision of generated motions. For instance, when generating a smile, the model may also learn to produce eyebrow-lowering actions, which could conflict with user-specified eyebrow motions (\eg brow raise). To enhance accuracy, it is essential to decouple the generation of different facial regions. However, certain facial motion couplings are crucial for conveying naturalness—for instance, a subtle head movement that accompanies a shift in gaze. Therefore, the model must achieve a delicate balance, selectively decoupling motions to improve accuracy while preserving necessary couplings to enhance naturalness.



\noindent\textbf{Generation Model $\mathcal{G}$.}
$\mathcal{G}$ is a diffusion-based model. 
Diffusion models consist of two Markov chains~\cite{ho2020ddpm,song2020score}: the forward chain incrementally injects Gaussian noise into the original signal, while the reverse chain sequentially reconstructs the original signal from the noise. $\mathcal{G}$ predicts the original signal instead of noise, and the loss function is introduced as:
\begin{equation}
	\mathcal{L}_{\eqword{denoise}} = \mathbb{E}_{t\sim \mathcal{U}[1, T],~\vect{M}_{(0)},\vect{C} }(\|\vect{M}_{(0)}-\mathcal{G}(\vect{M}_{(t)}, t, \vect{C})\|^2),
\end{equation}
where $t$ denotes the diffusion step, $\vect{M}_{(0)}$ is the original facial motion sequence, and $\vect{M}_{(t)}$ is the noisy sequence produced by the diffusion forward function $q(\vect{M}_{(t)} | \vect{M}_{(t-1)}) = \mathcal{N}(\sqrt{\alpha_n}\vect{M}_{(t-1)}, (1-\alpha_n)\vect{I})$. $\vect{C}$ is the timeline condition.

To balance decoupling and coupling for both accuracy and naturalness,
the generation model employs a base-branch design (\cref{fig:generation_pipeline}). It consists of a base network and individual branch networks for each facial region (\eg upper face, lower face). The base network takes timelines of all facial regions and noisy motions as input and encodes global motion couplings into base features.
The branch network takes the base features the timelines relevant only to its designated facial region as inputs, and generates decoupled motions for each region.

We use the the expression, eye, and pose coefficients of FaceVerse~\cite{wang2022faceverse} 3DMM model as target motion representation for generation. Its expression coefficients align with ARKit blendshapes, allowing separate representations of different facial regions.
When dividing facial motions into different regions for separate generations, we observe that splitting them into the upper face, lower face, and pose \& other regions strikes a better balance between decoupling and necessary coupling, rather than assigning a branch network for each individual region (\eg further split eye and brow). 
Specifically, the upper face includes eye, brow, and gaze; the lower face includes mouth and jaw; and the pose \& other regions cover the head pose and remaining regions like cheek and nose. It is important to note that, since pose is coupled with facial motions in all regions, the timeline control for the pose \& other branch includes all facial regions (rather than just the control for pose).

As for the specific implementation, the timeline control for each frame $\vect{c}_i$ is implemented as a 16-dimensional vector composed of values 0 and 1, with each dimension representing an action. A value of 1 indicates that the action is performed in that frame, and a value of 0 indicates that the action is not present. The timeline control is transformed into timeline tokens through a linear projection.
The base network and branch networks share the same structure. Their inputs consist of timeline tokens and motion tokens. 
The motion tokens are derived either from the noisy motion through linear projection or from base features. The motion tokens learn the temporal control from the timeline tokens through cross-attention in multiple transformer encoder layers. 
When incorporated into each encoder layer, the timeline tokens always use the initial timeline tokens rather than the output tokens from the previous layer. This prevents the temporal information in timeline tokens from being altered, thereby enhancing motion accuracy.
The motion tokens for each layer come from the output of the motion tokens from the previous layer. The network's final output consists of the motion tokens from the last layer, which are then linearly projected to produce the facial motion. 
Each branch network generates the facial motion for its corresponding region. These motions are then combined to produce the motion for the entire face.

To improve flexibility and generalization, we use classifier-free guidance~\cite{ho2022classifierfree} to train our model. During training, we randomly drop the condition of each facial region timeline. we use a dropout probability of 0.5 for each condition independently, with a 0.1 probability of dropping all conditions and a 0.1 probability of maintaining all conditions. When the timeline condition for a certain region is dropped, its value is set to -1.





\noindent\textbf{Rendering.}
We develop a diffusion-based portrait animation method to render motions as images. Since the renderer is not the main contribution of this paper, its details are reported in the \supp.

\subsection{Translating Natural Language to Timelines}
We observe that ChatGPT can generate a reasonable timeline from natural language descriptions, enabling our model to produce facial motions based on natural language input. The generated timeline can be further edited by users to fit their needs. To enhance the realism of the generated movements, we manually annotated several timeline descriptions as examples, allowing ChatGPT to perform few-shot learning.
The details of the prompts and generated results are reported in \supp.

%% file: sec/4_experiment.tex
\section{Experiments}


\noindent\textbf{Dataset.}
To effectively model natural facial motions, a dataset capturing authentic facial expressions is essential.
For this purpose, we utilize the RealTalk dataset~\cite{geng2023affective}. 
The RealTalk dataset contains 692 videos captured from genuine, unscripted conversations, resulting in highly natural facial movements with rich temporal dynamics.
Previously used lab-recorded, scripted datasets, such as MEAD~\cite{wang2020mead}, lack these qualities.
We use 1412 video clips extracted from Realtalk as the dataset. The size of the dataset is approximately 600,000 frames.  



\subsection{Facial Motion Annotation}
\label{sec:exp_facial_motion_annotation}
\noindent\textbf{Baselines.}
No previous methods have attempted annotating intervals of facial actions in a labor-efficient manner. We experimented with time series analysis methods such as AutoPlait~\cite{matsubara2014autoplait}, CubeMarker~\cite{honda2019multi}, and TICC. However, AutoPlait and CubeMarker failed to produce satisfactory results, as the motions of the segments within each cluster were inconsistent, making annotation impossible. Therefore, we primarily relied on TICC. 
Empirical results~\cite{hallac2017toeplitz} indicate TICC’s annotation performance can be enhanced by tuning key hyperparameters, including the number of clusters and beta, a hyperparameter that facilitates temporal consistency. We examine the effects of these parameters.



\noindent\textbf{Evaluation Metric.}
Annotating intervals of facial actions is essentially a multi-class classification problem, where each data point represents a frame and each class represents an action.
To evaluate annotation accuracy, we manually annotate the facial actions on 50 videos to serve as ground truth, and use Macro-F1 as the metric for evaluation. For each class, the F1 score is the harmonic mean of the precision and recall of our estimate. Then, the macro-F1 score is the average of the F1 scores for all the classes. 
A higher Macro-F1 score indicates greater annotation accuracy. 
We calculated the macro-F1 score for the eye region only on segments when the eyes are not closed.

\begin{figure}[t!]
  \centering
  \includegraphics[width=0.47\textwidth]{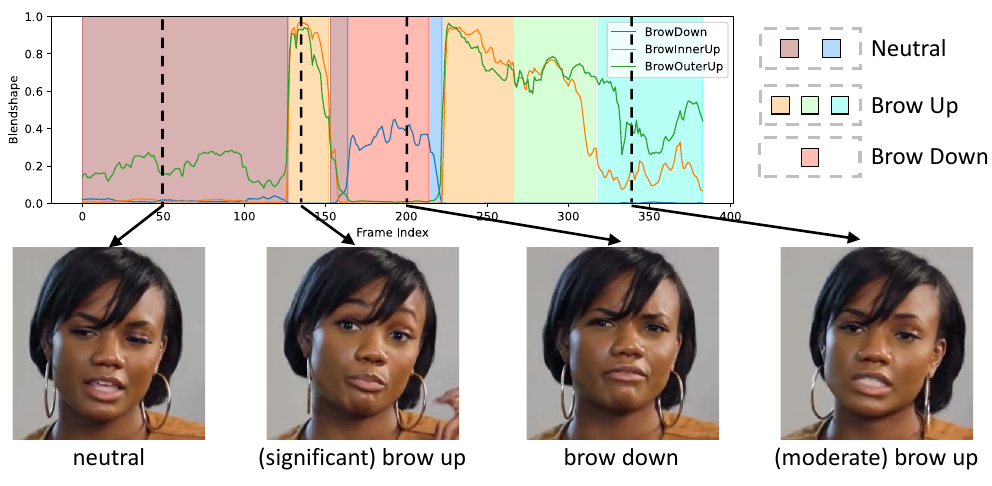}
  \vspace{-3mm}
  \caption{An example of brow motion annotation.}
  \vspace{-2mm}
  \label{fig:brow_motion_annotation}
\end{figure}

\begin{figure}[t!]
  \centering
  \includegraphics[width=0.47\textwidth]{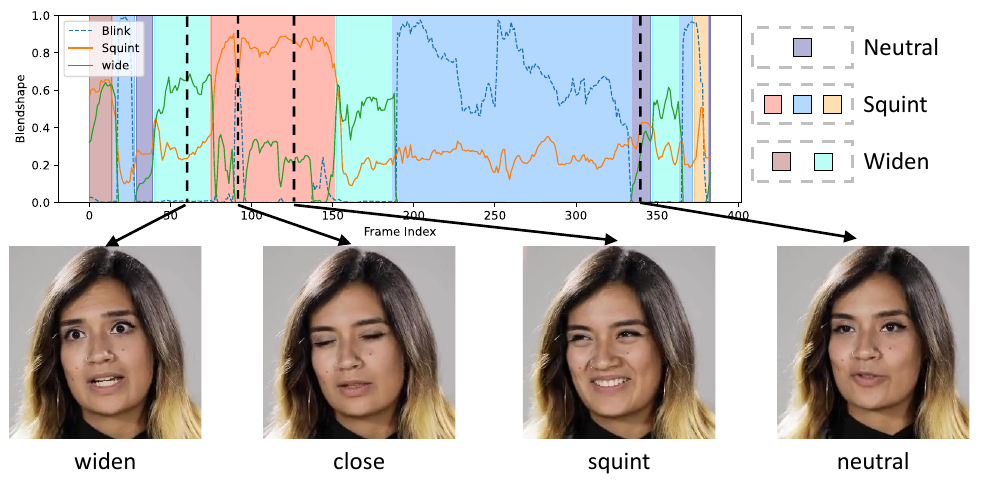}
  \vspace{-3mm}
  \caption{An example of eye motion annotation.}
  \vspace{-2mm}
  \label{fig:eye_motion_annotation}
\end{figure}

\noindent\textbf{Qualitative results.} \cref{fig:brow_motion_annotation} shows brow motion annotation. Our method can precisely segment different facial motion patterns and cluster similar ones. For a specific facial motion brow raise, our approach can distinguish multiple variants, such as a significant or moderate raise. In this work, these finer categories are grouped under one motion category for simplicity, but future methods could generate each subcategory separately. Annotation for the mouth is similar to that for the brow. \cref{fig:eye_motion_annotation} shows eye motion annotation. As stated in \cref{sec:method_facial_motion_annotation}, only eye squint/widen are annotated using TICC. Eye closure is determined by a threshold on the blendshape \emph{eyeBlink} and may occur within intervals of other eye motions. We also conduct a user study to evaluate annotation accuracy and the results are reported in \cref{sec:exp_facial_motion_generation}.

\begin{figure}[t!]
  \centering
  \includegraphics[width=0.47\textwidth]{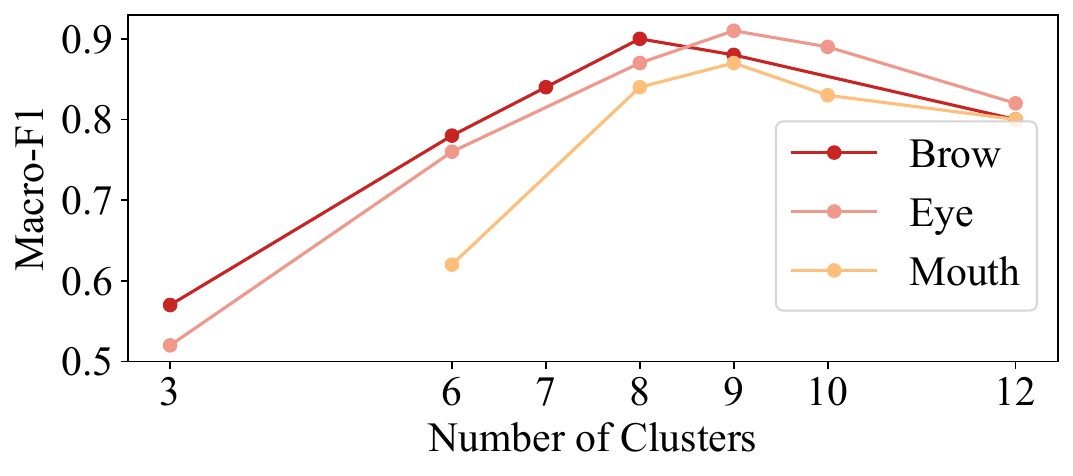}
  \vspace{-3mm}
  \caption{The annotation accuracy for different number of clusters.
  }
  \vspace{-2mm}
  \label{fig:cluster_num_impact}
\end{figure}

\begin{figure}[t!]
  \centering
  \includegraphics[width=0.47\textwidth]{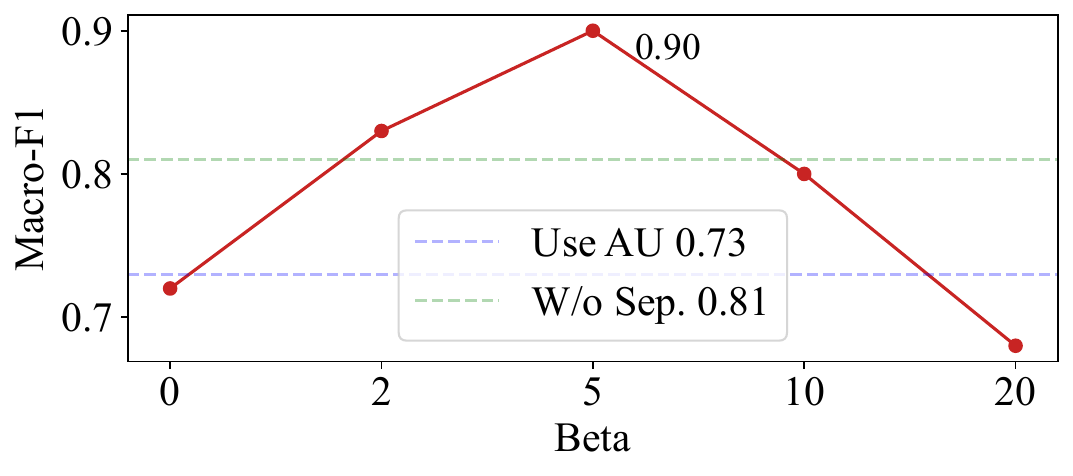}
  \vspace{-3mm}
  \caption{The annotation accuracy in brows for different $\beta$, concatenating videos without separating, and using AU as descriptors.
  }
  \vspace{-2mm}
  \label{fig:beta_impact}
\end{figure}

\noindent\textbf{Quantitative results.}
We select the optimal hyperparameters based on the highest macro-F1 score.
Experiments show that the optimal beta value for different regions is 5, with the best number of clusters being 8 for the eye region, 9 for the brow region, and 9 for the mouth region. (The number of clusters here refers to the count of valid clusters, excluding the one that contain only null sequences). The optimal scores for brow, eyes, and mouth are 0.90, 0.91, and 0.87, respectively. When using threshold-based annotation, the scores for eye closure, pose, and gaze are 0.95, 0.87, and 0.89, respectively.

The impact of different cluster numbers is shown in \cref{fig:cluster_num_impact} (with beta set to 5). Too few clusters in TICC lead to under-segmentation, grouping distinct patterns into single clusters.
Conversely, too many clusters decreases inter-cluster variability, making clusters less distinct
and blurring boundaries.

\cref{fig:beta_impact} illustrates the impact of $\beta$, using brow as an example (with the number of clusters set to 8). $\beta$ controls the smoothness of segment transitions.  Too large $\beta$ leads to over-smoothing, 
Conversely, a small $\beta$ makes the model overly sensitive, leading to  noisy segments that misrepresent the data’s structure.
\cref{fig:beta_impact} shows that when using AUs, the score decreases to 0.73. This highlights the importance of accurate facial motion descriptor quality for annotation.   \cref{fig:beta_impact} also shows that separating different video clips when analyzing improves performance.

To conduct a preliminary test of TICC's effectiveness in detecting unseen facial motion sequences, we remove 50 manually annotated videos from the dataset, refit TICC, and then have it annotate these videos to calculate the macro-F1 score. The scores are similar to those obtained when the 50 videos are included (brow: 0.89; eyes: 0.88; mouth: 0.85).


\begin{figure}[t!]
  \centering
  \includegraphics[width=0.47\textwidth]{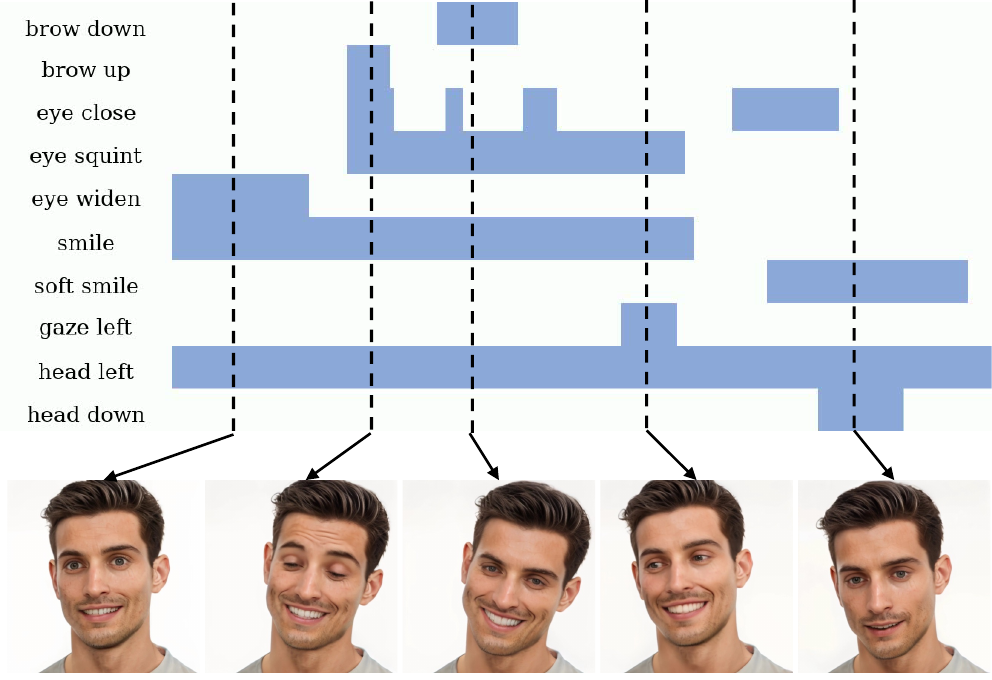}
  \vspace{-3mm}
  \caption{Qualitative results of facial motion generation from the timeline. Better viewed in \suppvideo. The timelines for the unplotted regions are set to 0. The same applies to the subsequent figures.}
  \vspace{-2mm}
  \label{fig:facial_motion_generation_qualitative_result_A}
\end{figure}

\begin{figure}[t!]
  \centering
  \includegraphics[width=0.47\textwidth]{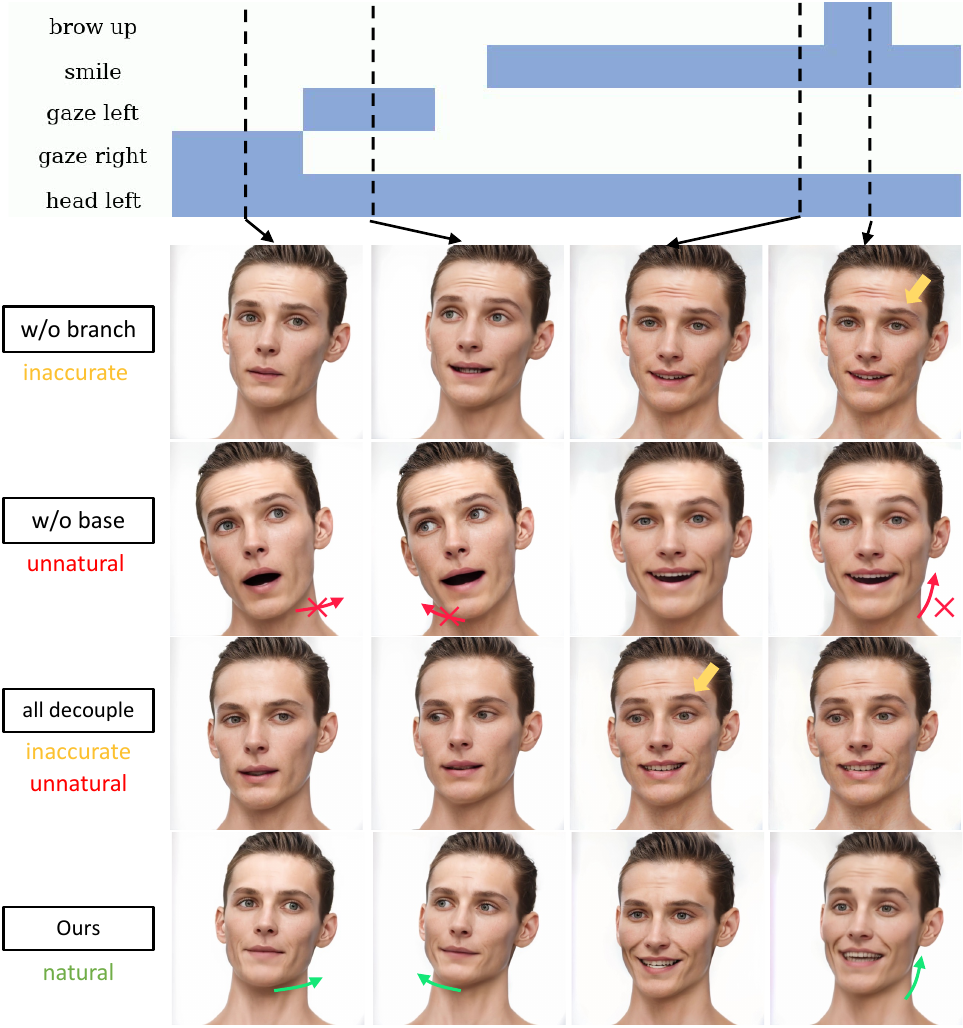}
  \vspace{-3mm}
  \caption{Qualitative results of ablation study. Green arrows indicate subtle natural head motions, red arrows indicate missing subtle natural head motions, and yellow arrows highlight inaccurate motions. Better viewed in \suppvideo.}
  \vspace{-2mm}
  \label{fig:ablation}
\end{figure}

\begin{table}[t]
    \centering
    \setlength{\tabcolsep}{0.5mm}{
    \begin{tabular}{cccccc}
    Methods & Var$\rightarrow$ & $\text{FID}_{fm}$ $\downarrow$ & $\text{FID}_{\Delta fm}$ $\downarrow$ & SND $\downarrow$ & TAS $\uparrow$ \\ 
        \shline
        w/o branch & 0.68 & 7.39 & 0.14 & 7.53 & 0.66  \\ 
        w/o base & 0.64 & 12.4 & 0.18 & 12.58 & 0.81  \\ 
        all decoup. & 0.41 & 28.4 & 0.23 & 28.63 & 0.69  \\ 
        \hline
        w/o time con. & \textbf{0.71} & 5.38 & 0.10 & 5.48 & 0.79  \\ 
        branchL1 & \underline{0.70} & 6.25 & 0.11 & 6.36 & 0.76  \\ 
        branchL3 & 0.68 & 5.76 & 0.10 & 5.86 & \textbf{0.84}  \\ 
        branchL4 & 0.69 & 6.64 & 0.12 & 6.76 & 0.83  \\ 
        \hline
        drop 0 & 0.62 & 6.88 & 0.13 & 7.01 & 0.78  \\ 
        drop 0.3 & 0.67 & 5.93 & 0.11 & 6.04 & 0.83  \\ 
        drop 0.7 & 0.78 & \textbf{4.12} & \textbf{0.09} & \textbf{4.21} & 0.68  \\ 
        \hline
        \textbf{Ours} & \underline{0.70} & \underline{4.54} & \textbf{0.09} & \underline{4.63} & \textbf{0.84} \\ 

    \end{tabular}}
    \caption{Results of ablation study. 
    The unit of \(\text{Var}\) in the table is \(10^{-2}\), meaning that \(0.70\) in the table represents \(0.70 \times 10^{-2}\). The unit of $\text{FID}_{fm}$ and $\text{FID}_{\Delta fm}$ is \(10^{-1}\). Our results differ in magnitude from those of ~\cite{yu2023talking} because we use a different 3DMM (FaceVerse) for the calculations.
    "$\rightarrow$" means results are better if they are close to the variance of real data, which is $0.73$.}
    \label{tab:ablation}
\end{table}

\subsection{Facial Motion Generation}
\label{sec:exp_facial_motion_generation}

\noindent\textbf{Evaluation Metric.} 
We utilize 100 timelines unseen during training to generate samples for evaluating the motion accuracy and naturalness. To evaluate accuracy, we use TICC or threshold-based approach to annotate facial motion intervals within the generated video. These intervals are then compared to the input timeline to calculate the macro-F1 score for each facial region. The average of these scores across all regions is used to assess the alignment between the generated motion and the input timeline.  We refer to this score as Timeline Alignment Score (\textbf{TAS}). To evaluate naturalness, we follow previous methods~\cite{yu2022talking,wang2023agentavatar} and use \textbf{Var} (variance of generated facial motions, with values closer to GT indicating better performance), $\textbf{FID}_{fm}$ (FID score of 3DMM coefficients),  $\textbf{FID}_{\Delta fm}$ (FID score of 3DMM coefficient difference between consecutive frames), \textbf{SND} (sum of $\textbf{FID}_{fm}$ and $\textbf{FID}_{\Delta fm}$).

\noindent\textbf{Qualitative Results.}
\cref{fig:facial_motion_generation_qualitative_result_A} shows the results generated by our model. Our model is capable of producing natural facial movements that are aligned with the input timeline. Note that because the training data includes videos of people speaking, our generated videos sometimes include speaking. 
Since FaceVerse 3DMM decouples the motion of the left and right face, we can generate motion separately for each side, allowing for generating asymmetric expressions.
Our method can generate forehead wrinkles when the speaker raises brows.
By introducing descriptors that capture more complex texture-related motions, we can extend our method to control them.

\noindent\textbf{Quantitative Results \& Ablation Study.}
To evaluate the impact of our design choices, we conduct an ablation study with several variants: (1) \textbf{w/o branch} remove all branches and only use the base network to predict the facial motions; (2) \textbf{w/o base} remove the base network and only use branch networks to generate each region independently; (3) \textbf{all decoup.} decouple all facial and head motions into different branch networks rather than limiting the decoupling to the upper and lower halves of the face. (4) \textbf{w/o time con.} starting from the second layer of the base/branch network, the initial timeline token input is replaced with the timeline tokens from the previous layer.
We investigate the optimal number of layers for the branch network. The entire network consists of 8 layers, with \textbf{branchL$x$} indicating that the branch network uses $x$ layers while the base network uses $8-x$ layers. In the optimal version \textbf{Ours}, the branch network has 2 layers.
We investigate the optimal drop probability. \textbf{drop $x$} indicates the drop probability is $x$, the optimal probability used in \textbf{Ours} is $0.5$

\cref{tab:ablation} shows quantitative results and \cref{fig:ablation} shows qualitative results from a timeline. 
\textbf{w/o branch} fails to generate accurate motions, highlighting the necessity of using the branch network to enhance accuracy.
When the head is oriented to the left, subtle movements may occur due to changes in gaze and brow. While these subtle motions do not affect the overall head direction, they are essential for motion naturalness. \textbf{w/o base} fails to generate these subtle motions (\cref{fig:ablation}).
Even though the pose branch receives the complete motion timeline, it still cannot produce such coupling. This highlights the necessity of using the base network to learn the coupling of facial movements. 
\textbf{all decoup.} produces stiff and inaccurate motion changes (brow raised too early in \cref{fig:ablation}) due to overfitting to the movements of individual regions, resulting in weak generalization to timelines outside the training set.
The performance of \textbf{w/o time con.} indicates that adding the initial timeline token at each layer can improve accuracy.
\textbf{branchL$x$}'s results show that increasing the number of branch layers excessively reduces naturalness. This is because having too few base layers is insufficient to model the natural coupling between motions.
\textbf{drop $x$}'s results show that too low or too high drop probability both cannot generate good performance. Too low probability makes the model overly rely on the condition and hampers its generalization capability.
Too high probability makes the model receive too few conditioning signals, reducing its accuracy.
\textbf{drop 0.7} get better FID scores, but its accuracy is low.

\begin{figure}[t!]
  \centering
  \includegraphics[width=0.5\textwidth]{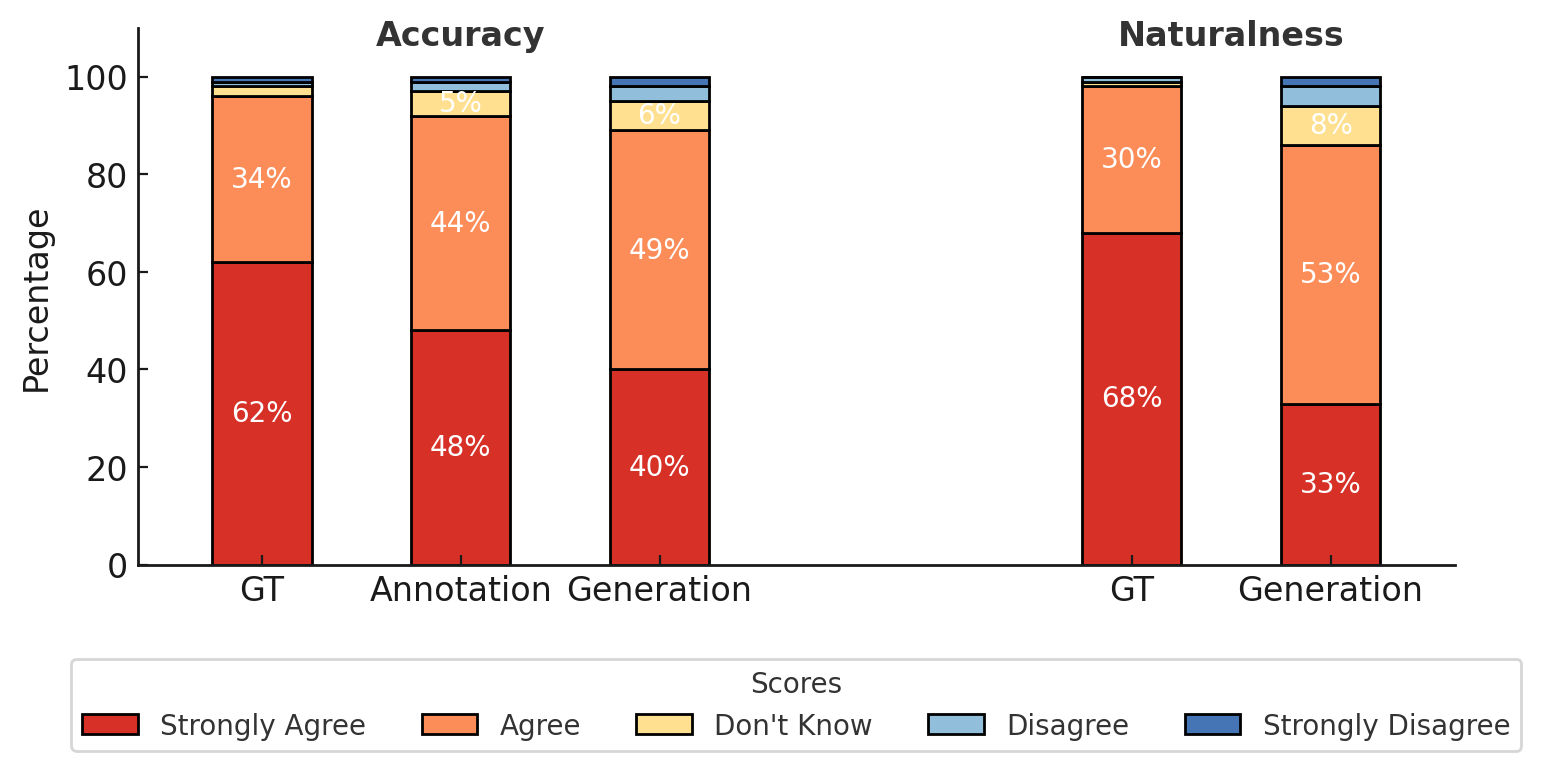}
  \vspace{-5mm}
  \caption{User study results.}
  \vspace{-8mm}
  \label{fig:user_study}
\end{figure}

\noindent\textbf{Comparisons with previous methods.} 
No prior methods have achieved fine-grained timeline control of facial motion generation. Similar approaches, such as AgentAvatar~\cite{wang2023agentavatar} and InstructAvatar~\cite{wang2024instructavatar}, can only describe temporal changes coarsely using temporal adverbs. Neither of these methods provides open-source code. We can only conduct a qualitative comparison using their demos. Since the temporal dynamics of facial motion are better demonstrated through video, these comparisons are included in \suppvideo.



\noindent\textbf{User Study.} We conduct a user study of 21 participants. The participants are required to rate three types of pairs: (1) \textbf{GT}: Real videos paired with manually annotated facial motion timelines. (2) \textbf{Annotation}: Real videos paired with timelines generated using our method. (3) \textbf{Generation}:  Videos generated by our method paired with the input facial motion timelines.
Participants need to evaluate their level of agreement (on a 5-point scale: strongly agree/disagree, agree/disagree, don't know) on two aspects:
(1) Whether the motions in the video \textbf{accurately} match the corresponding timelines.
(2) Whether the motions in the video appear \textbf{natural}. Each participant evaluates 15 pairs sampled from the test data for each type of pair, resulting in a total of 45 pairs per participant. \cref{fig:ablation} shows the results.  
92\% of the evaluations agree that our annotation is accurate. 89\% and 86\% of the evaluations consider the motions in our generated videos to be accurate and natural, respectively.

\begin{figure}[t!]
  \centering
  \includegraphics[width=0.47\textwidth]{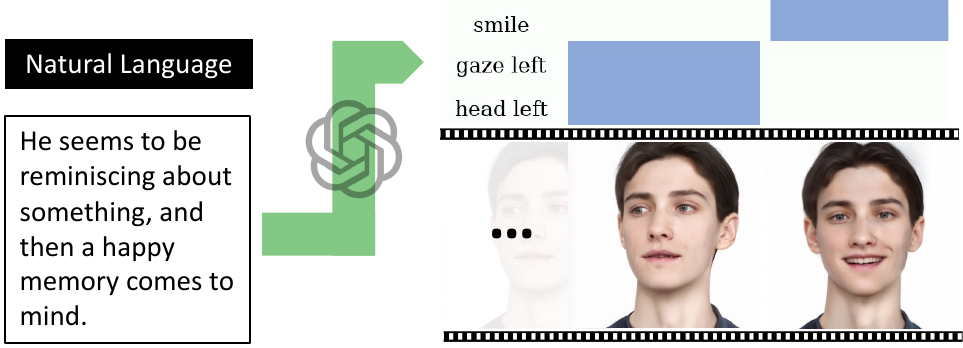}
  \vspace{-2mm}
  \caption{Our method supports text-guided facial motion generation by leveraging ChatGPT to convert text into timelines.}
  \vspace{-2mm}
  \label{fig:gpt}
\end{figure}

\noindent\textbf{Generating Motions Using Natural Language.}
Our method supports text-guided facial motion generation by leveraging ChatGPT to convert text into timelines. As shown in \cref{fig:gpt}, ChatGPT can infer that when a person is reminiscing, their head may turn to the side, and their gaze shifts accordingly. When a happy memory comes to mind, the person begins to smile.

%% file: sec/5_conclusion.tex
\section{Conclusion}
In this paper, we explore a new control signal: timeline control for facial motion generation. Timeline allows for more fine-grained control than audio or text, enabling users to generate specific motions with precise timing. To model this capability, we first develop a labor-efficient approach to annotate the temporal intervals of facial actions. The annotation process relies on time series analysis of facial motion descriptors. Based on the annotations, we propose a generation model that can generate natural facial motions aligned with timelines. 
The generation model utilize a base-branch design to effectively manage motion couplings across different facial regions. 
Our method supports text-guided generation by using ChatGPT to translate text into timelines. Experiments validate the effectiveness of our method.